\documentclass[10pt,twocolumn,letterpaper]{article}

\usepackage{myiccv}
\newcommand{\tablestyle}[2]{\setlength{\tabcolsep}{#1}\renewcommand{\arraystretch}{#2}\centering\footnotesize}

\usepackage{pdfpages}

\usepackage[pagebackref=true,breaklinks=true,letterpaper=true,colorlinks,urlcolor = black, bookmarks=false]{hyperref}
\usepackage{cleveref} 

\usepackage[breaklinks=true,bookmarks=false]{hyperref}
\usepackage{cleveref} 
\iccvfinalcopy 
\graphicspath{{./images/}}

\def\iccvPaperID{7107} 

\ificcvfinal\fi

\begin{document}

\title{Interpolation-Aware Padding for 3D Sparse Convolutional Neural Networks}

\author{
  {Yu-Qi Yang}$^{1,2}$\qquad
  {Peng-Shuai Wang}$^{2}$\qquad
  {Yang Liu}$^{2}$  \\
  $^1${Tsinghua University}\qquad
  $^2${Microsoft Research Asia}  \\
  {\tt\small yangyq18@mails.tsinghua.edu.cn\qquad
            \{penwan,yangliu\}@microsoft.com}
}

\maketitle
\ificcvfinal\thispagestyle{empty}\fi

\begin{abstract}
Sparse voxel-based 3D convolutional neural networks (CNNs) are widely used for various 3D vision tasks.
Sparse voxel-based 3D CNNs create sparse non-empty voxels from the 3D input and perform 3D convolution operations on them only.
We propose a simple yet effective padding scheme --- \emph{interpolation-aware padding} to pad a few empty voxels adjacent to the non-empty voxels and involve them in the 3D CNN computation so that all neighboring voxels exist when computing point-wise features via the trilinear interpolation.
For fine-grained 3D vision tasks where point-wise features are essential, like semantic segmentation and 3D detection, our network achieves higher prediction accuracy than the existing networks using the nearest neighbor interpolation or the normalized trilinear interpolation with the zero-padding or the octree-padding scheme.
Through extensive comparisons on various 3D segmentation and detection tasks, we demonstrate the superiority of 3D sparse CNNs with our padding scheme in conjunction with feature interpolation. 
\end{abstract}

\section{Introduction} \label{sec:intro}

Effective 3D representations for 3D deep learning like voxels \cite{Brock2016,Wu2016}, point sets~\cite{Qi2016,PointCNN}, and polygonal meshes~\cite{Wang2018,kato2018renderer}, have been actively studied in recent years. 
Among them, voxel-based representations are natural extensions from 2D pixels and are compatible with regular-grid-based convolutional operations and suitable for fast GPU processing. 
However, due to the high cost of memory storage and CNN computation on dense 3D grids, dense-voxel-based 3D CNNs are limited to coarse resolution inputs like $32^3$ grids, and cannot handle and generate high-resolution 3D contents. 
To overcome this limitation,  sparse-voxel-based CNNs~\cite{Wang2017,Riegler2017,Graham2018,choy20194d} are proved to be a computational and memory-efficient solution, where only voxels around 3D shapes are created for storing feature channels. 
With regard to the prediction accuracy on 3D tasks, spare-voxel-based CNNs dominate several large-scale benchmarks, including ScanNet segmentation~\cite{dai2017scannet}, KITTI segmentation~\cite{behley2019iccv}, and ScanNet detection~\cite{qi2019deep}. 

\begin{figure}[t]
    \centering
    \begin{overpic}[width=1\columnwidth]{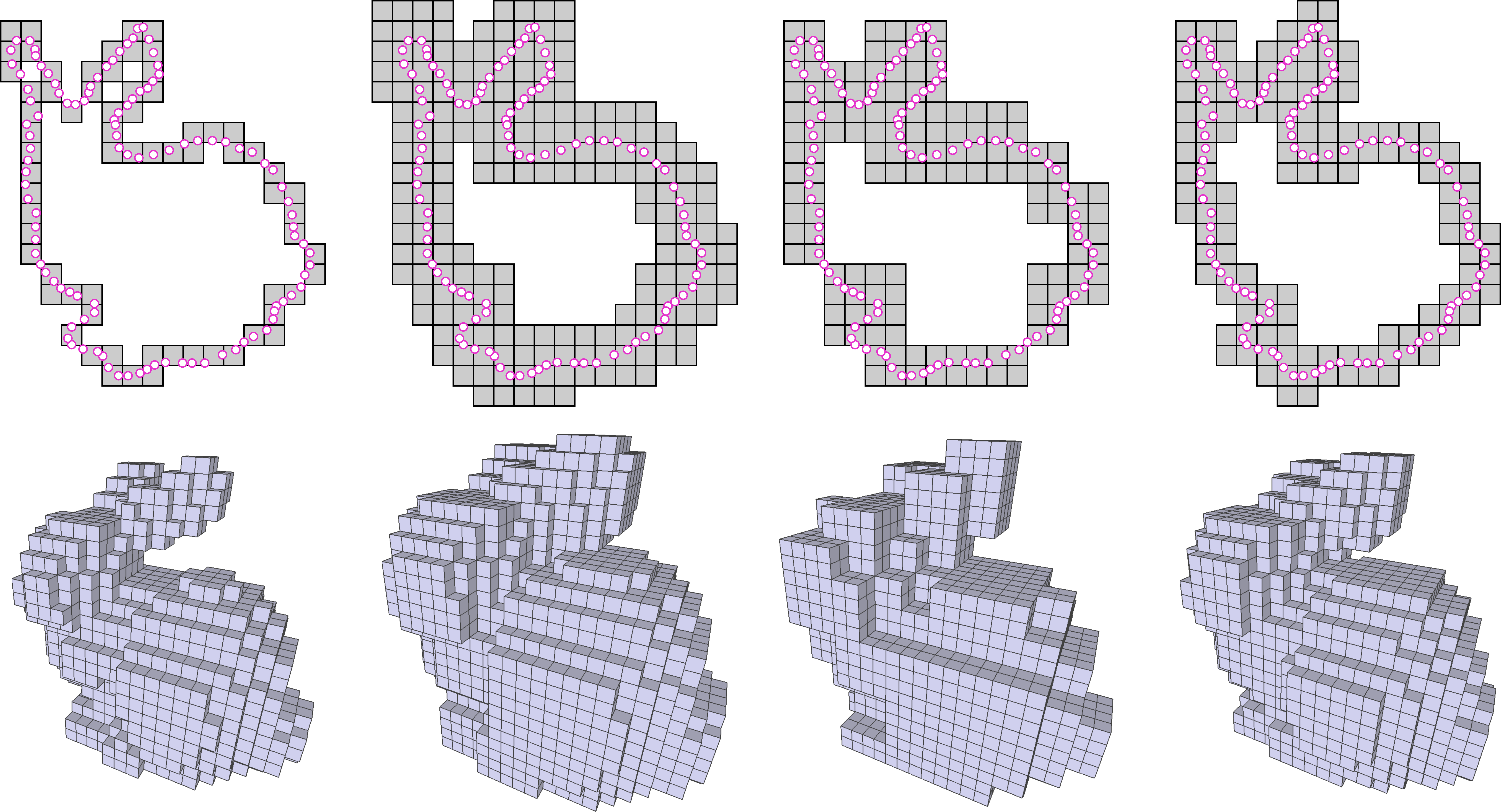}
        \put(10,-4){\small (a)}
        \put(36,-4){\small (b)}
        \put(62,-4){\small (c)}
        \put(88,-4){\small (d)}
    \end{overpic} \vspace{1pt}
    \caption{Various padding schemes for a 2D point set (upper) and a 3D point set (bottom). (a): sparse pixels/voxels without padding, \ie zero padding; (b): with 1-ring padding; (c): with quadtree-padding or octree-padding; (d): with our interpolation-aware padding. Red circles present 2D points sampled from a 2D bunny shape, 3D points are omitted here as they are hidden inside the voxels.
    } \label{fig:padding} 
\end{figure}
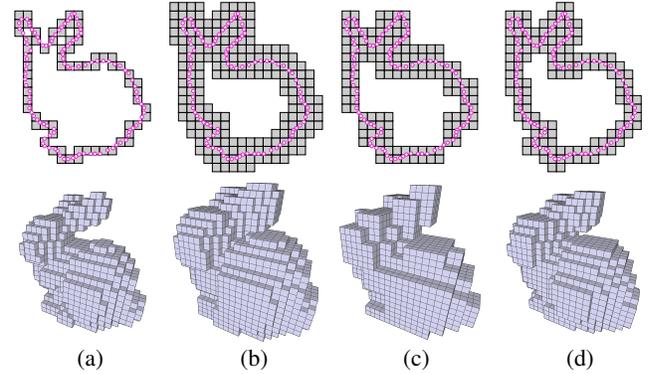

Sparse-voxel-based CNNs can be used naturally for extracting features for each discrete voxel. 
However, point-wise features are essential for fine-grained tasks, like 3D segmentation and detection. 
A common practice in sparse-voxel-based CNNs is to assign the feature of the nearest voxel to the point, \ie using \emph{the nearest neighbor interpolation}~\cite{Wang2017,Graham2018}. 
To extract distinguishable point features within the same voxel, \emph{voxel-based trilinear interpolation} can be adopted.
The trilinear interpolation requires querying features from the nearby eight voxels, whereas some nearby voxels may not exist due to the sparse-voxel representation.
Specifically, octree-based 3D CNNs~\cite{Wang2017,Wang2018a} perform the CNN computation on 8 sibling octants of non-empty octants (see Fig.~\ref{fig:padding}-(c)); 
the recent sparse-voxel-based CNNs~\cite{Graham2018,choy20194d,Shao2020} use Hash tables to index the non-empty voxels, and perform the CNN computation only in the non-empty voxels (see Fig.~\ref{fig:padding}-(a)).
To do the interpolation, existing works either assign zero features to those non-existing voxels~\cite{choy20194d}, or use normalized interpolation weights to compromising the missing voxels~\cite{Wang2020a,tang2020searching}.

We observe that current interpolation schemes are not optimal, and even yield worse performance than the nearest-neighbor interpolation in some experiments. 
Our key idea is to let the network learn features for the empty voxels, and use these learned features to do the interpolation, instead of using zero features.
To this end, a na\"{i}ve solution is to pad 1-ring neighbors of all original non-empty voxels when building the sparse voxel grids, denoted as \emph{1-ring padding} (see Fig.~\ref{fig:padding}-(b)).
Then these augmented voxels can be used as the input of CNNs, and all voxel features required by the interpolation can be computed by CNNs.
However, this may incur great computation and memory costs.
Considering that what we need are the point features, we can only pad those voxels required by the interpolation of each point (see Fig.~\ref{fig:padding}-(d)).
We term our padding scheme as \emph{interpolation-aware padding}. 
With our padding scheme, the interpolation is well-defined for each point, and the network performance can be greatly improved compared with the previous interpolation schemes. Compared with the 1-ring padding, the memory cost of interpolation-aware padding is also much less. \looseness=-1

Our improved interpolation also facilitates the network design.
Previous sparse-voxel-based CNNs for segmentation and detection output the features of discrete voxels~\cite{Wang2017,choy20194d}, the resolution of the output voxels has to be high enough to differentiate different points, otherwise the performance may decrease dramatically.
With well-defined interpolation, our network instead outputs \emph{coarse} resolution voxel features to extract expressive point-wise features, and further reduces the computation and memory cost.
Apart from the benefits of the interpolation, we also compare the four different padding schemes in Fig.~\ref{fig:padding}.
Interestingly, our experiments reveal that padding itself also boosts the network prediction accuracy, even without the trilinear interpolation.

Our contributions include the improved interpolation for sparse-voxel CNNs by the interpolation-aware padding scheme, and  network architectures for 3D segmentation and detection which can produce point-wise features without using very high-resolution voxels. To validate the efficacy and superiority of our method, we performed a series of experiments and comparisons on several typical 3D semantic segmentation and 3D detection tasks. 
Compared with the network using zero-padding and without interpolation, our network with the same number of trainable parameters improves the mean Intersection over Union (mIoU) of segmentation on PartNet~\cite{Mo2019}, ScanNet~\cite{dai2017scannet} and  KITTI~\cite{behley2019iccv} by 2.2, 2.0, and 2.4, respectively, and improves the mAP of detection on ScanNet~\cite{dai2017scannet} by 2.0. Compared with the 1-ring padding, our interpolation-aware padding scheme is more practical with less memory consumption. 
In the supplemental material, we provide our code for reproducing all the results easily. 
We believe our effective interpolation and sparse padding scheme will be a powerful plug-in for sparse 3D CNNs and benefit more broad applications.

\section{Related Work} \label{sec:related}

\paragraph{Dense voxel-based 3D CNNs}
Grid-based voxelization is a popular 3D discretization method in computer graphics and computer vision.
The dense voxels are natural extensions of 2D pixels and suitable for building 3D convolutional neural networks from them.
For 3D closed shapes, early works \cite{Maturana2015,Wu2015} represent them as indicator functions or distance fields on dense voxels and apply 3D CNNs for recognizing 3D objects.
Brock \etal \shortcite{Brock2016} create a voxel-based 3D variational autoencoder for synthesizing 3D shapes.
Choy \etal \shortcite{Choy2016} bring the recurrent neural network to voxel-based 3D decoders for inferring 3D shapes from multiview images.
Voxel-based 3D generative adversarial networks (GAN) further enhance the generated shape quality~\cite{Wu2016}.
However, in 3D, dense voxels occupy $\mathcal{O}(n^3)$ storage spaces and lead to the costly CNN computation.
In practice, dense voxel-based 3D CNNs  work on low-resolution grids only. 

\paragraph{Sparse voxel-based 3D CNNs}
The sparsity of 3D data can be utilized to improve the efficiency of 3D CNNs.
Observing that most of the voxels have duplicated features, Riegler \etal \shortcite{Riegler2017} use a hybrid grid-octree to build 3D CNNs to support 3D learning at high resolutions.
As a 3D surface is usually a 2D manifold or the collection of 2D manifold patches, it only occupies limited spaces in 3D.
By discretizing the shape surface into voxels only, the surface can be represented by a set of sparse voxels.
Wang \etal \shortcite{Wang2017} propose octree-based CNNs (O-CNN), where the  CNN computation only takes place in the non-empty octants and their sibling octants in different octree levels.
Submanifold sparse convolutional networks ~\cite{Graham2018} and MinkowskiNet \cite{choy20194d} further restrict the storage and CNN computation on the non-empty voxels only, and result in less memory occupation and computation costs.
The spatial-hashing-based CNN~\cite{Shao2020} improves the efficiency of the hashing table by avoiding hash collision and reducing memory overhead.
Instead of using as minimal as possible voxels in CNN, our work shows that padding empty voxels properly can improve the performance of sparse 3D CNNs, without enlarging network weight parameter sizes.

\begin{figure*}
  \centering
  \begin{overpic}[width=0.95\linewidth]{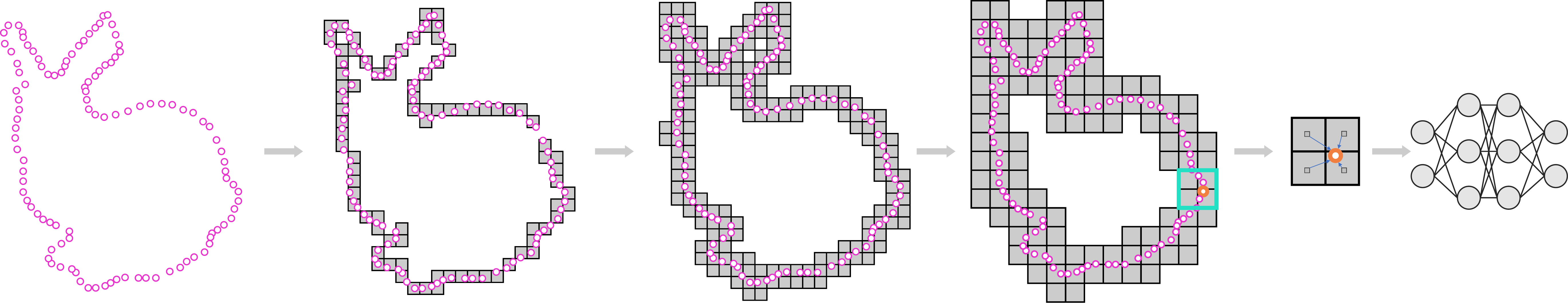}
    \put(2, -2) {\footnotesize (a) Input point cloud}
    \put(23,-2) {\footnotesize (b) Voxelization}
    \put(40,-2) {\footnotesize (c) Interpolation-aware padding}
    \put(65,-2) {\footnotesize (d) CNN features}
    \put(80,-2) {\footnotesize (e) Interpolation}
    \put(94,-2) {\footnotesize (f) MLP}
  \end{overpic}
  \vspace{10pt}
  \caption{Overview of our interpolation-aware padding scheme (2D illustration). 
  The input point cloud (a) is first voxelized (b) based on a given spatial resolution, and after voxelization, each of the voxel contains at least 1 point.
  Then the sparse non-empty voxels are padded so that all neighboring voxels used for interpolating each point exist (c).
  The sparse voxels are processed via deep sparse-voxel-based CNNs, and CNN features (d) are produced.
  Note that the spatial resolution of output CNN features (d) may be coarser than the input (b).
  For the orange point highlighted by a green box in (d), its neighboring voxels are retrieved, and the point feature is computed via interpolation (e).
  Finally, the point-wise features are used as the input of a shared MLP (f) for segmentation and detection tasks.} 
  \label{fig:overview} 
\end{figure*}
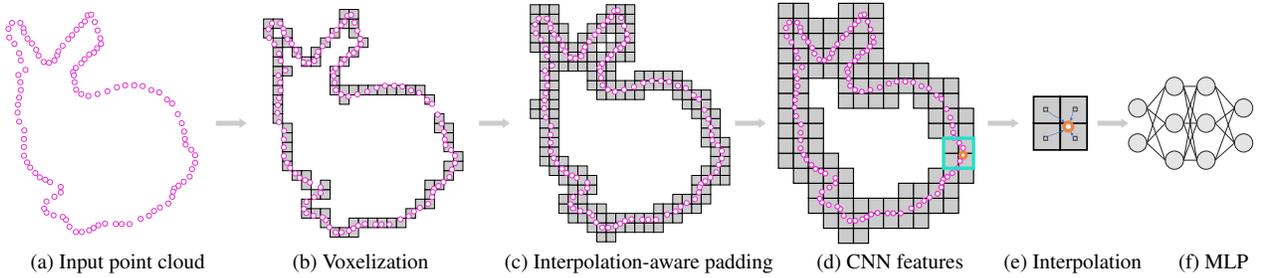

\paragraph{Sparse convolution}
The 3D sparse convolution widely used for 3D signals~\cite{Wang2017,Graham2018,choy20194d} is essentially a restricted version of a dense convolution, where the input feature maps on the non-existing voxels are set to zero during the convolution computation.
The convolution operation can be efficiently implemented by the GPU-version sparse matrix multiplication.
The convolution results can be further normalized by the number of non-zero features to achieve better performance for depth completion and image inpainting tasks~\cite{Uhrig2017,partialconv,partialconv1,Huang2020}.
In our work, we simply use the unnormalized sparse convolution.

\paragraph{Trilinear feature interpolation on sparse voxels}
For tasks requiring point-wise features, the voxel features produced by sparse-voxel-based CNNs can be scattered to each point via trilinear interpolation.
Trilinear interpolation on sparse voxels is not thoroughly evaluated and two different implementations spread in literature and the code repositories.
Since not all neighboring voxels exist, Choy \etal \shortcite{choy20194d} propose to use the zero feature directly if the corresponding voxel does not exist.
Wang \etal \shortcite{Wang2020a} and Tang \etal \shortcite{tang2020searching} use the normalized interpolation weights to satisfy the partition of unity property in their implementation.
With our interpolation-aware padding, we can directly use the original interpolation formula.
Mao \etal \shortcite{mao2019iccv} propose a new point-based convolution operation based on trilinear interpolation.
Here we focus on point-wise feature extraction for sparse-voxel-based CNNs and omit the comparison with~\cite{mao2019iccv}.
\section{Method} \label{sec:method}

Given an input point cloud, we train a sparse-voxel-based CNN to extract point-wise features for the downstream tasks like 3D segmentation and detection.
The overall pipeline is shown in Fig.~\ref{fig:overview}.
The input point cloud is first quantized and rounded to sparse voxels with a user-specified spatial resolution.
Instead of directly performing sparse convolutions on these non-empty voxels~\cite{Graham2018,choy20194d,Shao2020}, we pad a set of empty voxels required by the interpolation for each point and set the initial input features of these empty voxels as zero.
After a set of CNN operations, each of the voxels contains its extracted features.
For a query 3D point, its eight nearest neighboring voxels are retrieved, and the voxel features are interpolated to produce its point-wise feature. Then the point-wise feature is processed via two fully-connected layers and used for each specific task.

In the following sections, we first introduce our novel interpolation-aware sparse padding and the trilinear interpolation in Sec.~\ref{sec:padding}, network designs in Sec.~\ref{sec:network}.

\subsection{Interpolation-aware Sparse Padding} \label{sec:padding}

The padding for sparse-voxel-based CNNs here is fundamentally different from the conventional 2D image padding. 
We pad extra voxels around the empty voxels as the input.
Initially, the input features of the padded voxels are set as zero, and then all voxel features are dynamically produced by the CNN operations.
For the 2D convolution on images, \emph{virtually}-padded pixels are needed for pixels close to the image boundary due to the fixed shape of convolution kernels.
And the features of the padded pixels are directly set to zeros (\ie zero padding) or the feature maps at the reflection pixels on the image boundary (\ie reflection padding). 

Our goal is to pad extra voxels around the non-empty voxels so that all the eight nearest neighboring voxels exist when doing the trilinear interpolation for each point. 
The trilinear interpolation inside a 3D regular grid ~\cite{wiki:trilinear} can be written as:
\begin{equation}
    \mathbf{f}(x,y,z) := \dfrac{\sum I_{ijk} \cdot \vol\nolimits_{ijk} \cdot \mathbf{f}_{ijk}}{\sum I_{ijk} \cdot \vol\nolimits_{ijk}}, 
    \label{eq:interpolation}
\end{equation}
where $i,j,k \in \{0, 1\}$ are the indices of the eight grid corners,  $\vol_{ijk}$ is the partial volume surrounded the by the query point and the corner which is diagonally opposite to the corner with index $ijk$, $\mathbf{f}_{ijk}$ is the feature vector associated on corner $ijk$, and $I_{ijk} \in \{0, 1\}$ is indicates whether the corresponding voxel exists.
In the following, we examine two baseline padding schemes and propose our interpolation-aware padding scheme for balancing the network performance and the runtime memory cost.

\paragraph{Octree padding}
The work of O-CNN~\cite{Wang2017} explicitly builds sparse voxels using octrees. 
The octree padding is an outcome of the octree data structure since each octree node has 8 children nodes.
Fig.~\ref{fig:padding}-(c) illustrates an octree padding and its 2D counterpart.
However, the eight nearest neighboring voxels used for interpolation of each input point are still not guaranteed to exist.
We regard this padding scheme as one of the baselines for comparison.

\paragraph{$N$-ring padding}
The most direct way of padding voxels for interpolation is to add all the empty voxels whose Manhattan Distance from their centers to the center of one of the non-empty voxels is not greater than $N$, where $N$ is a positive integer.
In this way, all the voxels used in the Eq.~\ref{eq:interpolation} exist.
Fig.~\ref{fig:padding}-(b) illustrates $1$-ring padding in 2D and 3D.

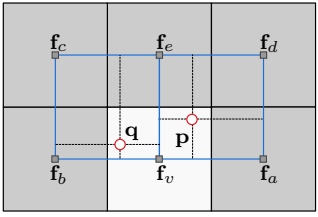
\begin{figure}[t]
    \centering
    \begin{overpic}[width=0.5\columnwidth]{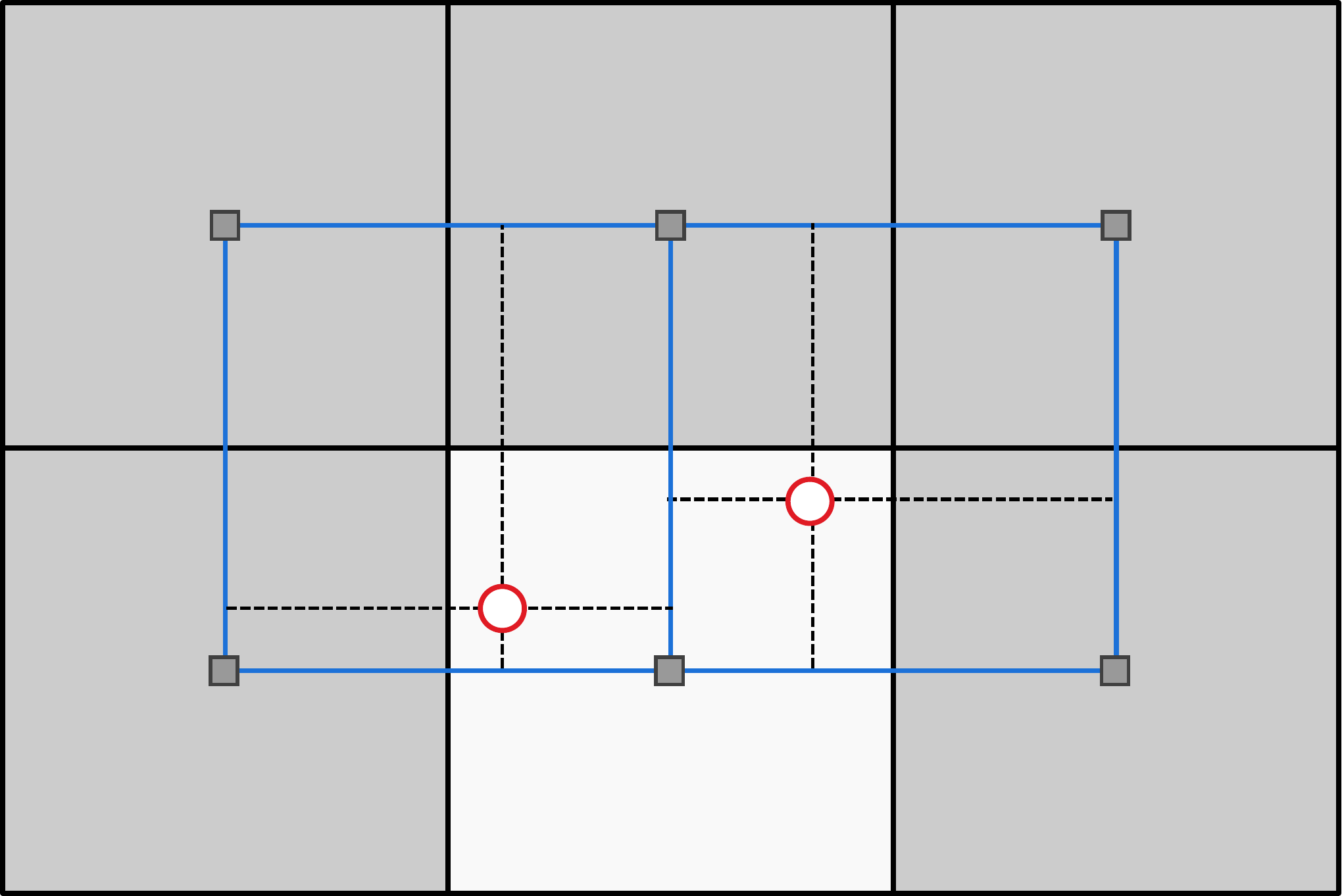}
        \put(55,22){\footnotesize $\mathbf{p}$}
        \put(39,24){\footnotesize $\mathbf{q}$}
        \put(82,10){\footnotesize $\mathbf{f}_a$}
        \put(15,10){\footnotesize $\mathbf{f}_b$}
        \put(15,52){\footnotesize $\mathbf{f}_c$}
        \put(82,52){\footnotesize $\mathbf{f}_d$}
        \put(49,52){\footnotesize $\mathbf{f}_e$}
        \put(49,10){\footnotesize $\mathbf{f}_v$}
    \end{overpic}
    \caption{2D illustration of interpolation-aware padding. 
    Two points $\mathbf{p}$ and $\mathbf{q}$ are inside the light gray pixel. 
    Because the bilinear interpolation at these two points queries the CNN features $\mathbf{f}_a,\mathbf{f}_b,\mathbf{f}_c,\mathbf{f}_d,\mathbf{f}_e,\mathbf{f}_v$, dark gray pixels are explicitly added into the sparse pixel set.}
    \label{fig:interpolation} 
     \vspace{-3mm}
\end{figure}

\paragraph{Interpolation-aware padding}
To interpolate CNN features on a 3D point $\mathbf{p}=(x,y,z)$ which is inside a non-empty voxel $\mathbf{v}$ using Eq.~\eqref{eq:interpolation}, we pad all the empty voxels involved in the interpolation into the CNN computation. 
We assume the 3D bounding box of the input is specified and the corner with the minimal x-, y-, z- coordinates is denoted by $\mathbf{p}_o = (x_o,y_o,z_o)$.
The voxel indices of all the eight voxels involved by the interpolation are computed as follows:
\begin{equation*}
    \resizebox{.95\linewidth}{!}{$
            I_x := \lfloor \frac{x-x_o}{s} + o_x \rfloor, \; I_y := \lfloor \frac{y-y_o}{s} + o_y \rfloor, \; I_z := \lfloor \frac{z-z_o}{s} + o_z \rfloor,
        $}
\end{equation*}
where $\lfloor \cdot \rfloor$ is the floor function, $s$ is the voxel size (\ie voxel edge length), and $o_x, o_y, o_z \in \{0.5, -0.5\}$.  
The center locations of these voxels are $ \mathbf{p}_o + s (I_x + 0.5, I_y + 0.5, I_z + 0.5). $
Fig.~\ref{fig:interpolation} shows a 2D illustration of interpolation-aware padding where two 2D points appear inside pixel $v$. It can be seen that the number of padded pixels/voxels depend on the point location, which can vary from $0$ to $26$ in 3D. 
Due to the construction of interpolation-aware padding, the set of padded voxels is a subset of voxels of 1-ring padding.

\paragraph{Complexity statistics of sparse padding}
We denote $M$ as the non-empty voxel number, the worst-case memory and computational complexity of octree padding, 1-ring padding, and our interpolation-aware padding are $\mathcal{O}(8 \cdot M)$, $\mathcal{O}(27 \cdot M)$, and $\mathcal{O}(27 \cdot M)$, respectively.
The worst case happens only when the average point spacing of the input point cloud is larger than $2s$, \ie all the non-empty voxels are disjointed with each other. 
For interpolation-aware padding, the worst-case further requires that eight corner regions of every non-empty voxel contain points, and the padding appears in all directions.
The former extreme situation may exist in using high-resolution sparse voxels for the point cloud from LiDAR scans, and the latter case for interpolation-aware padding does not happen on real inputs. 
For point sets from 3D objects and indoor scenes, the commonly-used sparse voxel size is bigger, less memory cost and computation efforts can be achieved.

\begin{figure}
    \centering
    \begin{overpic}[width=\columnwidth]{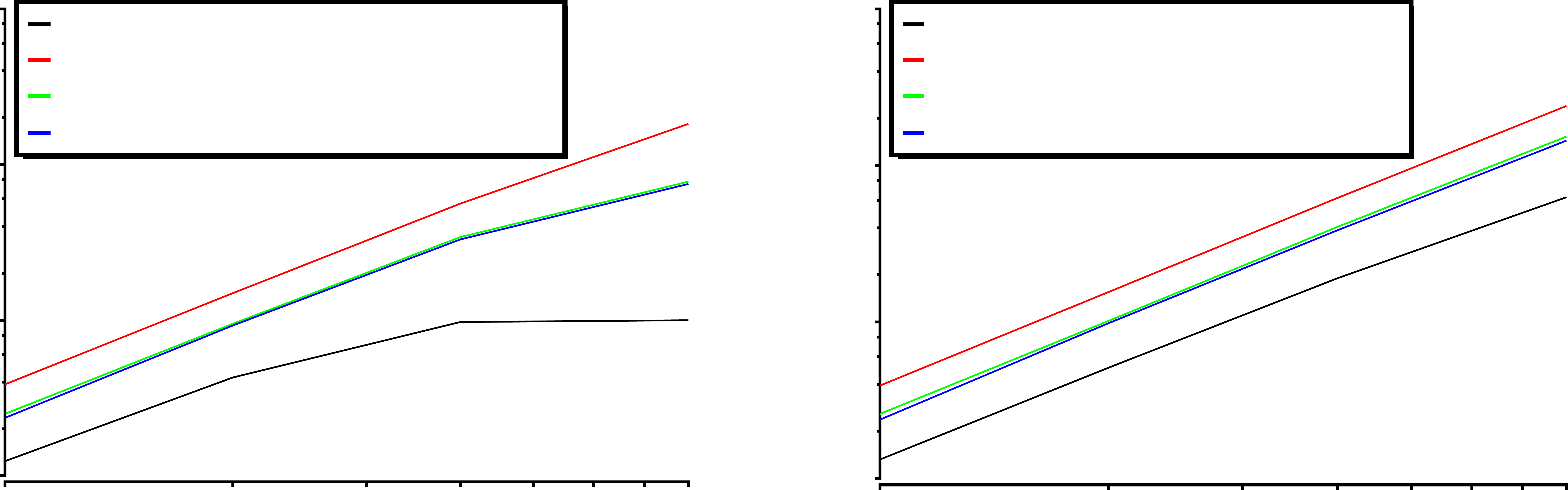}
        \put(5,29.2){\fontsize{5}{6}\selectfont zero padding}
        \put(5,26.8){\fontsize{5}{6}\selectfont 1-ring padding}
        \put(5,24.5){\fontsize{5}{6}\selectfont octree padding}
        \put(5,22.1){\fontsize{5}{6}\selectfont interpolation-aware padding}
        \put(60,29.2){\fontsize{5}{6}\selectfont zero padding}
        \put(60,26.8){\fontsize{5}{6}\selectfont 1-ring padding}
        \put(60,24.5){\fontsize{5}{6}\selectfont octree padding}
        \put(60,22.1){\fontsize{5}{6}\selectfont interpolation-aware padding}
        \put(-5, 33){\scriptsize \# voxels}
        \put(51, 33){\scriptsize \# voxels}
        \put(50,30){\fontsize{5}{6}\selectfont $10^6$}
        \put(50,20){\fontsize{5}{6}\selectfont $10^5$}
        \put(50,10){\fontsize{5}{6}\selectfont $10^4$}
        \put(50,0){\fontsize{5}{6}\selectfont $10^3$}
        \put(-6,30){\fontsize{5}{6}\selectfont $10^6$}
        \put(-6,20){\fontsize{5}{6}\selectfont $10^5$}
        \put(-6,10){\fontsize{5}{6}\selectfont $10^4$}
        \put(-6,0){\fontsize{5}{6}\selectfont $10^3$}
        \put(-2,-3){\fontsize{5}{6}\selectfont $\frac{1}{10}$}
        \put(13,-3){\fontsize{5}{6}\selectfont $\frac{1}{20}$}
        \put(27,-3){\fontsize{5}{6}\selectfont $\frac{1}{40}$}
        \put(41,-3){\fontsize{5}{6}\selectfont $\frac{1}{80}$}
        \put(44,1){\scriptsize $s$}
        \put(54,-3){\fontsize{5}{6}\selectfont $\frac{1}{10}$}
        \put(68,-3){\fontsize{5}{6}\selectfont $\frac{1}{20}$}
        \put(83,-3){\fontsize{5}{6}\selectfont $\frac{1}{40}$}
        \put(97,-3){\fontsize{5}{6}\selectfont $\frac{1}{80}$}
        \put(101,1){\scriptsize $s$}
    \end{overpic}
    \vspace{1mm}
    \caption{Statistics of the number of occupied voxels by different padding schemes under different voxel sizes. The voxel size $s$ is selected from $\{\frac{1}{10}, \frac{1}{20}, \frac{1}{40}, \frac{1}{80} \}$. The bounding box size of the input Bunny model is $2$.  The numbers of the input points are $10000$ (left) and $100000$ (right), respectively. } \label{fig:plot}\vspace{-3mm}
\end{figure}

In Fig.~\ref{fig:plot}, we calculate the total voxel number after applying different padding schemes on an example, whose input points are uniformly sampled from a 3D Bunny model. 
The number of padded voxels by interpolation-aware padding is quite similar to the one by octree padding, which is about $6.8$ times the number of non-empty voxels when the voxels are highly sparse (10000 points with $s=1/80$); while it is just $1.4$ times when the sparsity is moderate.

\begin{figure*}[h!]
    \centering
    \begin{overpic}[width=0.95\linewidth]{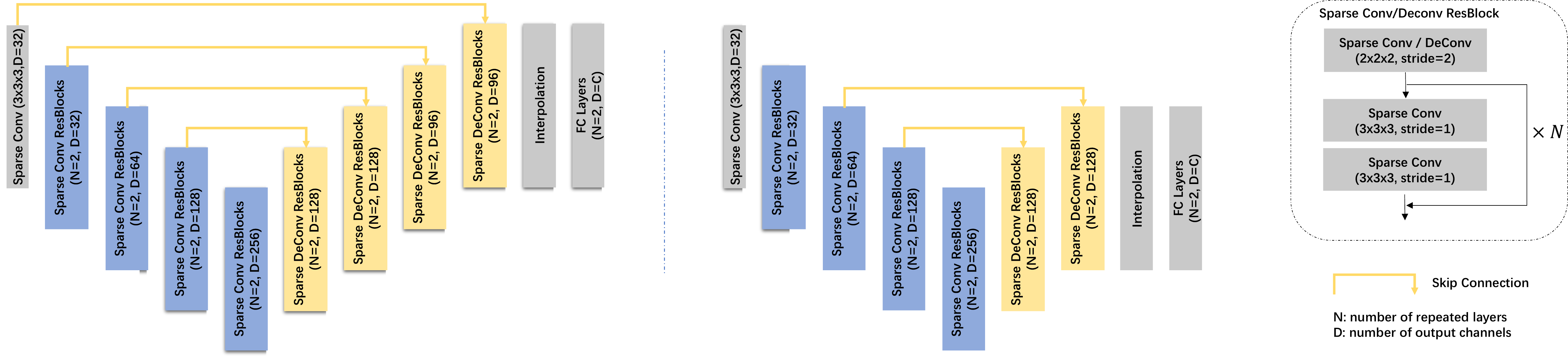}
    \end{overpic}
    \caption{Sparse-voxel based U-Net structure.
    On the left panel, the resolution of output voxels is the same as the input; on the right panel, the resolution of output voxels is coarser.
    }
    \label{fig:net} \vspace{-3mm}
\end{figure*}

\subsection{Network design} \label{sec:network}
We use the U-Net structure with five levels of domain resolution as shown in the left panel of Fig.~\ref{fig:net}, which consists of multiple sparse-convolution-based residual blocks and skip connections. 
The point feature is interpolated from the output voxel features at the finest level. 
Two additional FC layers are appended after the interpolated feature. 

The network can also interpolate point-wise features from coarse-resolution voxel features.
An example is shown in the right panel of Fig.~\ref{fig:net}.
In the decoder of U-Net, the ResNet blocks operating on the two highest resolutions are dropped, which may further reduce the computation and memory cost while maintaining similar performances.
Compared with U-Net network with the nearest neighbor interpolation, the output voxel feature has to be at the finest resolution, otherwise, multiple points in the same voxel can not be distinguished, which may harm the performance.

The proposed interpolation-aware padding can be added to all resolution levels of the sparse voxel grids, which achieves the best performance according to our experiments.
However, according to the complexity analysis in the previous section, the padding on the finest resolution consumes the most resources.
So we add the padding to the level corresponding to the output voxel resolution only for reducing computation and memory cost.

\section{Experimental Analysis} \label{sec:result}

In this section, we choose the fine-grained part segmentation task on the PartNet dataset to evaluate our interpolation and padding schemes, and the network structures, with highlights on our contributions. 
By default, all experiments were tested on a Linux server with Intel Core I7-6850K CPU (\SI{3.6}{GHz}) and a GeForce GTX 2080 Ti GPU (\SI{11}{GB} memory). The implementation is built upon the sparse voxel CNNs provided by~\cite{tang2020searching}.

\paragraph{Dataset} 
We pick four categories (Chair, Lamp, Storage furniture, Table) of PartNet as the benchmark, each of which has at least 1000 shapes and contains three levels of semantic labels, and we follow the original data splitting for training and test.
The input data (10000 points) is normalized to fit inside a unit box, and the finest grid resolution in the network is set to $64^3$. 
The network outputs the finest-grained semantic label (level-3) at points.

\paragraph{Experiment setting}
We train the U-Net in Fig.~\ref{fig:net} with an output resolution $64^3$ to evaluate different interpolation and padding schemes.
We name the U-Net integrated with one of padding schemes: zero padding, 1-ring padding, octree padding, and interpolation-aware padding by \textsc{Zero}, \textsc{Ring}, \textsc{Octree}, and \textsc{Interp}, respectively.
And we denote the nearest neighbor and trilinear interpolation as \textsc{Near} and \textsc{Linear}.
We train the U-Net with one of the above settings for each shape category and evaluate the segmentation quality on the test data with the part mean IOU metric~\cite{Mo2019}. 
For a fair comparison, each network was trained three times and all the networks are initialized with the same set of parameters each time. 
We use the SGD optimizer with a learning rate of 0.1 and decay 0.1 at 1/2 and 3/4 of the max epoch. The batch size is set to 24.
We report the averaged metric with mean deviation in Tab.~\ref{tab:partnetseg}. The IoU metrics on individual categories are reported in the supplemental material. 

\begin{table}[t]
  \tablestyle{4pt}{1.1}
  \begin{tabular}{ccccccc}
    \toprule
    \thead{Group}         & \thead{Pad.}     & \thead{Interp.} & \thead{$\mathbf{S_{out}}$}  & \thead{mIoU}          & \thead{Time}  & \thead{Mem.}  \\
    \midrule
    \multirow{4}{*}{(1)}  & \textsc{Zero}    & \textsc{Near}   &$64^3$    & $40.5\pm0.2$          & 382           & 1471          \\
                          & \textsc{Octree}  & \textsc{Near}   &$64^3$    & $40.0\pm0.2$          & 568           & 2560          \\
                          & \textsc{Ring}    & \textsc{Near}   &$64^3$    & $\mathbf{40.9}\pm0.0$ & 929           & 3790          \\
                          & \textsc{Interp}  & \textsc{Near}   &$64^3$    & $40.6\pm0.1$          & 623           & 2707          \\
    \midrule
    \multirow{4}{*}{(2)} & \textsc{Zero}     & \textsc{Linear} &$64^3$  & $41.5\pm0.0$          & 398           & 1622          \\
                         & \textsc{Octree}   & \textsc{Linear} &$64^3$  & $41.4\pm0.0$          & 591           & 2610          \\
                         & \textsc{Ring}     & \textsc{Linear} &$64^3$  & $\mathbf{42.7}\pm0.3$ & 965           & 3744          \\
                         & \textsc{Interp}   & \textsc{Linear} &$64^3$  & $42.3\pm0.3$          & 651           & 2758          \\
    \midrule
    \multirow{2}{*}{(3)} & \textsc{Zero}     & \textsc{Near}   &$32^3$  & $38.1\pm0.2$          & 248           & 902          \\
                         & \textsc{Interp}   & \textsc{Linear} &$32^3$  & $\mathbf{40.1}\pm0.1$ & 366           & 1646          \\
    \bottomrule
  \end{tabular}
  \caption{Quality statistics of fine-grained segmentation on four PartNet
    categories under different settings. \emph{Pad.} is the sparse padding type,
    \emph{Int.} is the sparse interpolation type, \emph{mIoU} is the average
    part IoU, \emph{Time} is the average time in milliseconds of a single
    forward and backward propagation on a batch  (16 objects), and \emph{Mem.}
    is the average GPU memory (in Megabyte) occupied by a batch. The experiments
    are grouped for analysis.} \label{tab:partnetseg} \vspace{-4mm}
\end{table}

\paragraph{Performance improvement by the interpolation}
By comparing the results between Group (1) and (2) in Tab.~\ref{tab:partnetseg}, we can see that the performances with the trilinear interpolation (\textsc{Linear}) with different padding schemes are consistently better than the nearest neighbor interpolation (\textsc{Near}), which is understandable since different points inside one voxel can be distinguished via the trilinear interpolation. 

\begin{figure}[t]
  \centering
  \includegraphics[width=\columnwidth]{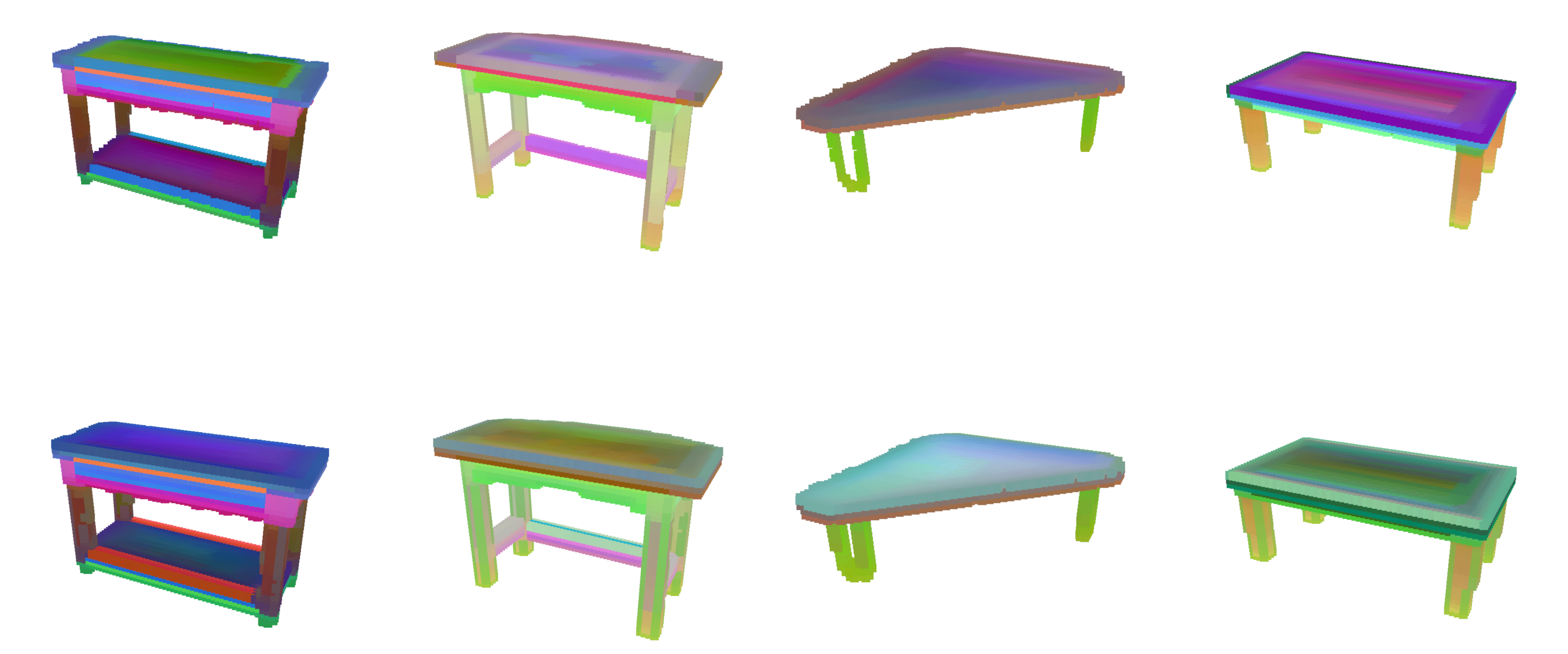}
  \caption{Visualization of learned features on non-empty voxels (top row) and padded voxels (bottom row).} \label{fig:features} 
\end{figure}

By comparing the padding schemes within Group (2), we can see that the U-Net with 1-ring padding (\textsc{Ring}) achieves the best performance, however, it consumes the largest GPU memory in the runtime and requires a longer execution time. 
The octree padding is no better than zero padding.
The U-Net with our interpolation-aware padding (\textsc{Interp}) and trilinear interpolation (\textsc{Linear}) achieves a good balance of accuracy gain and memory cost.
The interpolation-aware padding and 1-ring padding enable a well-defined trilinear interpolation and the padded voxels contain meaningful features after training, whereas zero features are used for empty voxels for octree padding and zero padding.
To verify that the learned features on padded voxels are different from zero or some constant, we visualize the output features of the U-Net with our interpolation-aware padding. 
We extract the output features on all the voxels and map them to the RGB domain via dimension reduction (T-SNE). 
Fig.~\ref{fig:features}-upper row shows the color map on the non-empty voxels, while Fig.~\ref{fig:features}-lower row shows the color map on the padded voxels. The features of the padded voxels are neither identical in a single object nor identical across different shapes.  

\begin{figure}[t]
  \centering
  \includegraphics[width=\columnwidth]{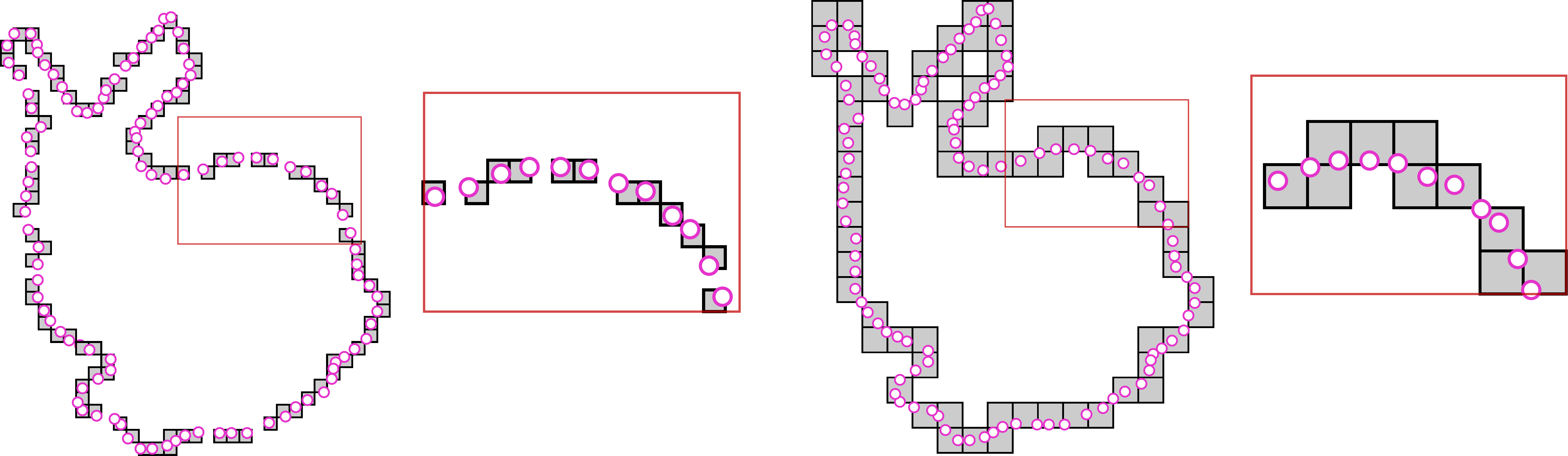}
  \caption{Non-empty pixels of a bunny shape under a high resolution (left) and a coarse resolution (right). 
  The sparse voxels are disjointed when the resolution is high.
  } \label{fig:nopadding}  \vspace{-3mm}
\end{figure}

\paragraph{The contributions of sparse padding}
By comparing the results within Group (1) in Tab.~\ref{tab:partnetseg}, we can see that the padding itself can also help to improve the performance, even without the trilinear interpolation.
We suspect that the reason is that padding around the non-empty voxels helps information propagation, especially for very sparse inputs.  
Fig.~\ref{fig:nopadding}-left illustrate  2D sparse non-empty pixels of a shape under a high resolution. 
For the disjointed non-empty pixels whose locations are close to each other (see the pixels around the gap region), the feature maps associated with them cannot be propagated to each other via any $3 \times 3$ convolution. 
Effective information propagation either requires a large-kernel-size convolution or occurs in coarse-version sparse pixels in which non-empty pixels have more neighbors as shown in Fig.~\ref{fig:nopadding}-right.
Although the worst complexity of interpolation-aware padding is $\mathcal{O}(27 \cdot M)$, in real scenarios like the PartNet experiments with the resolution $64^3$ the average complexity is $\mathcal{O}(1.95M)$ only, much lower than the worst case.

\paragraph{The improved network design}
We did experiments to validate the network which outputs coarse resolution features as shown on the right panel in Fig.~\ref{fig:net}.
The results with an output resolution of $32^3$ are reported in Group (3) in Tab.~\ref{tab:partnetseg}: 
compared with previous methods~\cite{Wang2017,Graham2018,choy20194d} with the zero-padding and the nearest neighbor interpolation, our performance is much better.
Compared with the network with an output resolution $64^3$ in Group (2), the memory and computation coast reduced by 40\%.
However, we notice that the overall mIoU is worse than Group (2), which is mainly caused by the Storage with an IoU decrease of 4.6 (more details can be found in the supplemental materials).
In Sec.~\ref{sec:cmps}, we also tested the U-Net outputting coarse resolution features in ScanNet segmentation (see Sec. ~\ref{subsec:scannet}) and KITTI segmentation (see Sec.~\ref{subsec:KITTI}), we find that in these experiments the mIoU is comparable or even better than the U-Net outputting fine resolution features.

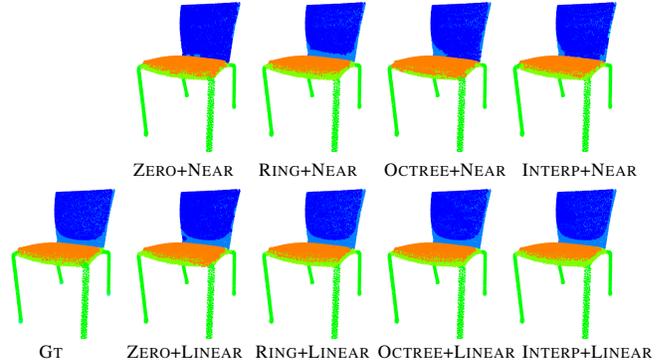
\begin{figure}[t]
  \centering
  \begin{overpic}[width=\columnwidth]{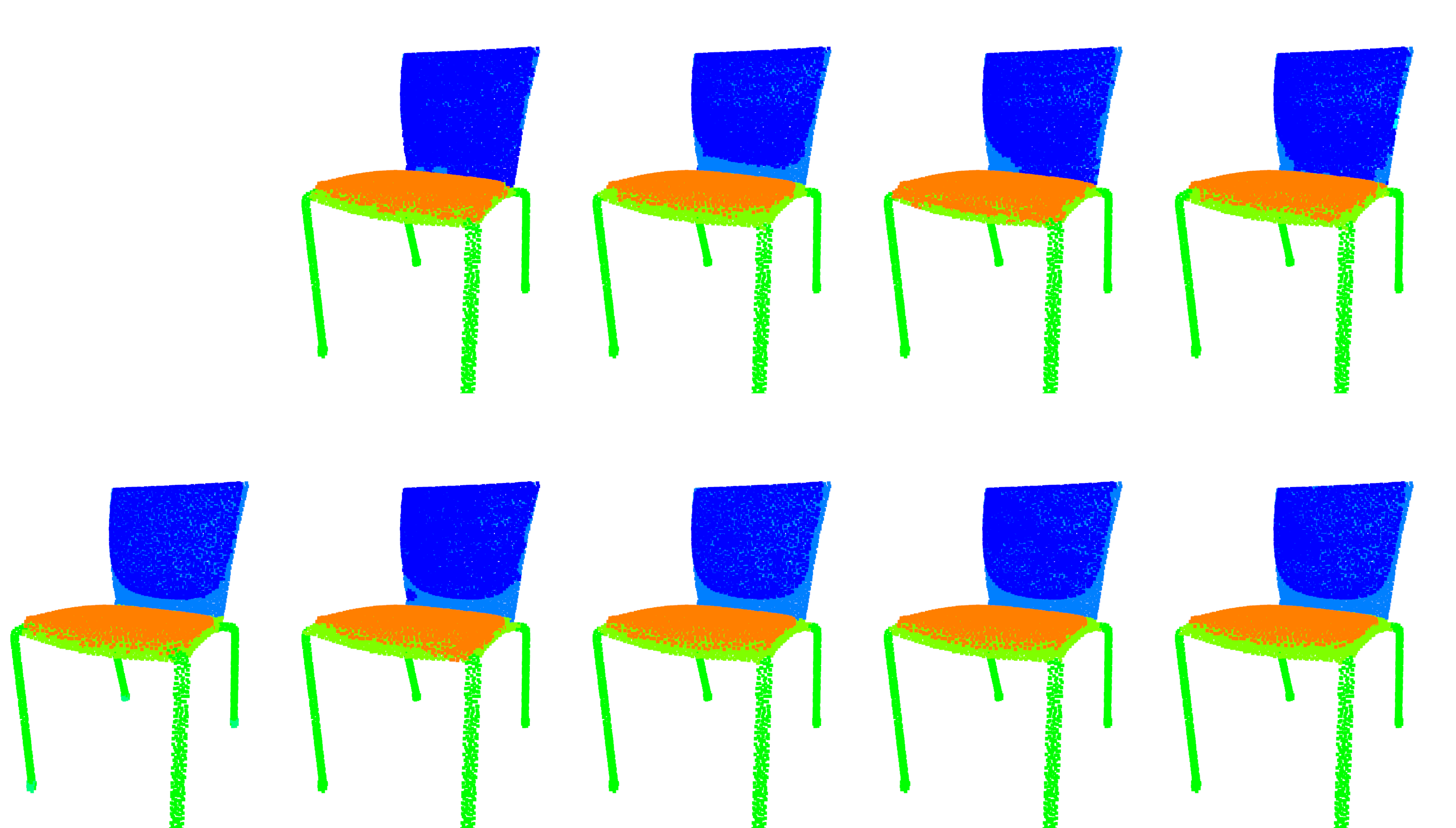}
  \put(20, 26) {\scriptsize \textsc{Zero}+\textsc{Near}}
  \put(40, 26) {\scriptsize \textsc{Ring}+\textsc{Near}}
  \put(60, 26) {\scriptsize \textsc{Octree}+\textsc{Near}}
  \put(82, 26) {\scriptsize \textsc{Interp}+\textsc{Near}}
  \put(5, -3)  {\scriptsize \textsc{Gt}}
  \put(19, -3) {\scriptsize \textsc{Zero}+\textsc{Linear}}
  \put(39.5, -3) {\scriptsize \textsc{Ring}+\textsc{Linear}}
  \put(59, -3) {\scriptsize \textsc{Octree}+\textsc{Linear}}
  \put(82, -3) {\scriptsize \textsc{Interp}+\textsc{Linear}}
  \end{overpic}
   \vspace{1pt}
  \caption{ Visualization comparison of segmentation results under different padding and interpolation schemes. Different colors indicate different part labels. } 
  \label{fig:vis} 
\end{figure}

\paragraph{Result visualization}
We visualize the segmentation results of a chair in the test dataset under different padding and interpolation schemes in Fig.~\ref{fig:vis}.  By comparing the results, we can see that the results with the trilinear interpolation (2nd row) are smoother and more faithful to the ground-truth than the results with the nearest interpolation (the first row).  
The results of \textsc{Ring} and \textsc{Interp} are always better than \textsc{Zero} and \textsc{Octree}. Zero padding has the worst segmentation results. \looseness=-1


\paragraph{Comparison with state-of-the-art}
In Tab.~\ref{tab:partnetseg_compare}, we report the comparison with other state-of-the-art methods, including PointNet++~\cite{qi2017pointnetplusplus}, PointCNN~\cite{PointCNN}, and O-CNN~\cite{Wang2020a}.
It can be seen that our results are much better than others in all four shape categories. 
This is mostly caused by the finer segmentation results at boundaries and fine-grained regions, as we can see from Fig.~\ref{fig:vis}. Our padding scheme improves the representation power of the interpolated point features and helps to distinguish segmentation region boundaries. 
Here we should mention that the employed O-CNN~\cite{Wang2020a} can be regarded as sparse-voxel-based CNNs with the trilinear interpolation and the octree padding in \emph{all} voxel resolutions.
For a fair comparison,  we also use our interpolation-aware padding in all voxel resolutions.
Note that the mIoU of octree padding and our interpolation-aware padding increases from 41.4 to 41.7, and 42.3 to 42.7 respectively, compared with the network with a single resolution padding as shown in Group (2) of Tab.~\ref{tab:partnetseg}.

\begin{table}[t]
    \tablestyle{6pt}{1.1}
    \begin{tabular}{lccccc}
      \toprule
      \thead{Method} & \thead{mIoU} & \thead{Chair} & \thead{Lamp} & \thead{Stora.} & \thead{Table}  \\
      \midrule
      PointNet++~\cite{qi2017pointnetplusplus}     & 34.7 & 39.2 & 25.3 & 40.5 & 33.9 \\
      PointCNN~\cite{PointCNN}                     & 33.7 & 43.9 & 20.1 & 49.4 & 21.3 \\
      O-CNN~\cite{Wang2020a}                       & 41.7 & 46.8 & 28.5 & 53.8 & 37.7 \\
      Ours                                          & \textbf{42.7} & \textbf{47.6} & \textbf{29.4} & \textbf{54.8} & \textbf{38.9} \\
      \bottomrule
    \end{tabular}
    \caption{Comparison with  state-of-the-art methods on PartNet.}
    \label{tab:partnetseg_compare} 
  \end{table}

\section{Comparisons}\label{sec:cmps}
We further evaluate and compare the performance of our method in other 3D segmentation and detection tasks with the state-of-the-art methods. Here we use the interpolation-aware padding for comparison, and the 1-ring padding is not tested as the network with 1-ring padding for on ScanNet and KITTI datasets runs out of memory even on a V100 GPU with \SI{32}{GB} memory. 

\subsection{Semantic segmentation on ScanNet} \label{subsec:scannet}

\paragraph{Dataset} 
The ScanNet dataset~\cite{dai2017scannet} contains 1.5k indoor scenes.
We follow \cite{choy20194d} to conduct the same data split and augmentation and feed the whole scene to the network without cropping.

\paragraph{Network structure}
We use the same U-Net with five levels of domain resolution as employed by MinkNet~\cite{choy20194d}. 
The network structure is same  to Fig.~\ref{fig:net}, but with more residual blocks (the numbers of repeated resblocks are [2,3,4,6,2,2,2,2]).
The only difference is that MinkNet uses the zero-padding and the nearest interpolation, while our network uses our interpolation-aware padding and trilinear interpolation.

\paragraph{Experiment setting}  
Similar to the setup of MinkNet~\cite{choy20194d}, the voxel size at the finest level in the encoder is set to \sicm{2} and the batch size is $9$. 
The input signal at each voxel is a 3-channel RGB color with 1 additional channel indicating whether the voxel is created by padding or not ($0$ for the padded voxels and $1$ for the original non-empty voxels).  
The voxel size used for point feature interpolation in the decoder is denoted by $S_{out}$, and we experiment different $S_{out}$ from \sicm{2} to \sicm{8} for evaluating the effect of feature interpolation.
For our network with feature interpolation, we only pad voxels at the level of $S_{out}$. 
The training scheme is the same as the one used in~\cite{choy2020high}: the optimizer is SGD and the learning rate is adjusted from 0.1 with the polynomial learning rate policy. We train all models for 600 epochs.
All the results are evaluated on the validation set of ScanNet, and are presented in Tab.~\ref{tab:scannetseg}.

\begin{table}[t]
  \tablestyle{4pt}{1.1}
  \begin{tabular}{lccccc}
    \toprule
    \thead{Network} & \thead{Pad.}  & \thead{Interp.} & $\mathbf{S_{out}}$ & \thead{mIOU} & \thead{Mem.} \\
    \midrule
    MinkNet \cite{choy20194d} &\textsc{Zero}         & \textsc{Near}       & \sicm{2}          & $72.2\pm0.3$ & 3514        \\
    Ours                       &\textsc{Interp}       & \textsc{Linear}       & \sicm{2}         & $\mathbf{72.8}\pm0.2$ & 6829        \\
    \midrule
    MinkNet \cite{choy20194d} &\textsc{Zero}         & \textsc{Near}       & \sicm{8}          & $70.4\pm0.1$ & 2184        \\
    Ours                       &\textsc{Interp}       & \textsc{Linear}       & \sicm{8}          & $\mathbf{72.4}\pm0.2$ & 2986        \\
    \bottomrule
  \end{tabular}
  \caption{ Quality statistics of semantic segmentation on ScanNet val set. 
  \emph{mIoU} is the average IOU of all the classes. }  \label{tab:scannetseg} 
\end{table}

\paragraph{Result analysis} 
The comparison in Tab~\ref{tab:scannetseg} further confirms the observations in Sec.~\ref{sec:result}: 
our network with feature interpolation achieves higher accuracy than MinkNet~\cite{choy20194d}. 
We also find that our network with feature interpolation at a coarser resolution (\sicm{8}) is comparable to MinkNet with high resolution (\sicm{2}) output: 72.4 vs. 72.2, while the runtime memory consumption is smaller: 2.9M vs. 3.5M.

\subsection{3D Object detection on ScanNet} \label{subsec:detection}

\paragraph{Dataset}
The ScanNet dataset contains instance segmentation labels of indoor scenes. 
With these labels, the bounding box of each object instance can be calculated. 
We follow the data preparation in the work of VoteNet~\cite{qi2019deep}, and use mAP@0.25 and mAP@0.5 as the evaluation metric.

\paragraph{Network structure}
We choose the original VoteNet as the baseline which uses PointNet++~\cite{qi2017pointnetplusplus} as the backbone to extract seed point features. 
We also replace PointNet++ with the 5-level U-Net mentioned in Sec.~\ref{subsec:scannet} and keep all other structures unchanged. 
The seed features are extracted from the third level of the decoder in the U-Net to mimic the features extracted from PointNet++'s SA2 layer which is used in the original VoteNet.

\paragraph{Experiment setting}
For U-Net, we set the voxel size of the finest level in the encoder to \sicm{2}, and the voxel size of the third level of the decoder $S_{out}$ is \sicm{8}. 
The batch size for all the networks is set to 8. 
Point color and height are the input signal, and we evaluate the detection results on the validation set.
The optimizer used in this task is Adam with a 0.001 initial learning rate. 
We train each model for 200 epochs.
The learning rate decays 0.3 at 80, 120, 160 epochs.

\paragraph{Result analysis}
With the stronger backbone --- sparse U-Net, the mAP@0.5 of object detection on ScanNet is increased by 3.2. 
And with the interpolated point feature and the interpolation-aware sparse padding, the performance (mAP@0.5) is increased by 6.7 from the baseline. 
Our network with the interpolation-aware sparse padding and the trilinear interpolation is also better than the zero-padding and the nearest neighbor interpolation, consistent with the previous segmentation experiments.  
Note that the sparse padding can be used as a plug-in in any method based on sparse voxel-based 3D convolution, we believe it can benefit other stronger baselines like 3D-MPA~\cite{Engelmann20CVPR}.

\begin{table}[t]
  \tablestyle{4pt}{1.1}
  \begin{tabular}{lcccc}
    \toprule
    \thead{Network}            & \thead{Pad.}    & \thead{Int}      & \thead{mAP@0.25}     & \thead{mAP@0.5}       \\
    \midrule
    VoteNet~\cite{qi2019deep}  & -               & -                & $57.8\pm0.6$         & $34.7\pm0.4$          \\
    MinkNet~\cite{choy20194d}  & \textsc{Zero}   & \textsc{Near}    & $58.7\pm0.5$         & $37.9\pm0.6$          \\
    Ours                        & \textsc{Interp} & \textsc{Linear}  & $\mathbf{60.7}\pm0.8$& $\mathbf{41.4}\pm0.6$ \\
    \bottomrule
  \end{tabular}
    \caption{ Quality statistics of instance detection on ScanNet validation set.
    VoteNet~\cite{qi2019deep} is based on PointNet++, we replace it with MinkNet~\cite{choy20194d} and report the results in the second row. The results of combining our interpolation-aware padding and interpolation scheme with MinkNet are shown in the last row. }  \label{tab:scannetdetection} 
\end{table}

\subsection{Semantic segmentation on KITTI Dataset} \label{subsec:KITTI}

\paragraph{Dataset}
Semantic KITTI~\cite{behley2019iccv} contains large-scale outdoor scenes annotated with semantic labels based on the 22 sequences~\cite{geiger2012cvpr}. 
Each sequence contains thousands of point clouds acquired by LiDAR sensors, in which the point density is quite non-uniform even in a single scan. We follow the standard train-validation split and report the results on the validation set.

\paragraph{Network structure}
We plug our interpolation-aware padding and trilinear interpolation into two networks on this task: the U-Net structure in Fig.~\ref{fig:net}, which is the same as the one used for MinkNet~\cite{choy20194d}, and the SPVCNN~\cite{tang2020searching}, which is composed of a low-resolution sparse-voxel-based U-Net and a high-resolution point-based branch.
Our padding and interpolation scheme is combined with the voxel-based U-Net part of SPVCNN.

\paragraph{Experiment setting}
We use the same setting of \cite{tang2020searching}: the voxel size of the finest level in the encoder is set to \sicm{5} and the batch size is set to $2$. 
The input signal contains a 3-channel point coordinate and a 1-channel LiDAR signal. 
Similar to the experiments in Sec.~\ref{subsec:scannet}, we experiment our padded-version networks with interpolated features from \sicm{5} and \sicm{20}.
The training scheme is the same as the one used in~\cite{choy20194d}: the optimizer is SGD and the learning rate starts from 0.24 and is adjusted by the cosine scheduler with warm-up. We train all models for 15 epochs.

\begin{table}[t]
  \tablestyle{4pt}{1.1}
  \begin{tabular}{lcccc}
    \toprule
    \thead{Network}              & \thead{Pad}        & \thead{Int}     & \thead{$\mathbf{S_{out}}$} & \thead{mIOU} \\
    \midrule
    MinkNet~\cite{choy20194d}    & \textsc{Zero}     & \textsc{Near}     & \sicm{5}          & $61.9\pm0.3$      \\
    MinkNet~\cite{choy20194d}    & \textsc{Zero}     & \textsc{Linear}   & \sicm{5}          & $61.4\pm0.1$      \\
    Ours                          & \textsc{Interp}    & \textsc{Linear}   & \sicm{5}          & $\mathbf{63.5}\pm0.3$ \\
    \midrule
    MinkNet~\cite{choy20194d}    & \textsc{Zero}     & \textsc{Near}     & \sicm{20}         & $61.5\pm0.3$            \\
    Ours                          & \textsc{Interp}    & \textsc{Linear}   & \sicm{20}         & $\mathbf{63.9}\pm0.4$ \\
    \specialrule{1pt}{1pt}{1pt}
    SPVCNN~\cite{tang2020searching} & \textsc{Zero}      & \textsc{Linear}   & \sicm{5}          & $62.9\pm0.7$       \\
    Ours                              & \textsc{Interp}    & \textsc{Linear}   & \sicm{5}          & $\mathbf{63.2}\pm0.3$      \\
    \midrule
    SPVCNN~\cite{tang2020searching} & \textsc{Zero}     & \textsc{Linear}   & \sicm{20}         & $62.8\pm0.2$          \\
    Ours                             & \textsc{Interp}   & \textsc{Linear}   & \sicm{20}         & $\mathbf{63.7}\pm0.1$   \\
    \bottomrule
  \end{tabular}
\caption{Quality statistics of semantic segmentation on KITTI dataset.
    In each panel between horizontal lines, \emph{Ours} means using the same network as upper lines, while adopting our interpolation-aware padding \textsc{Interp} and trilinear interpolation \textsc{Linear}.
    }  
\label{tab:kittiseg} 
\end{table}

\paragraph{Result analysis}
By using our interpolation-aware padding and trilinear interpolation as a plugin of MinkNet~\cite{Choy2016} and SPVCNN~\cite{tang2020searching}, we can see the performance increases consistently as shown in Tab.~\ref{tab:kittiseg}.
It is also interesting to find that our network with $S_{out}= \sicm{20}$ performs better than the settings with $S_{out}= \sicm{5}$. We speculate that the interpolation on coarser voxels may be suitable to handle extremely non-uniform distributed points. We also observe that under the zero padding setting, MinkNet with trilinear interpolation is not superior to the one with the nearest-neighbor interpolation, as shown in the first two rows of Tab.~\ref{tab:kittiseg}. The phenomenon shows that a proper padding scheme like our interpolation-aware padding is crucial for the trilinear interpolation. \looseness=-1

\section{Conclusion} \label{sec:conclusion}
In this work, we present an interpolation-aware padding that enables well-defined interpolation for sparse-voxel-based CNNs.
The efficacy of our padding schemes and improved networks is well demonstrated on 3D segmentation and 3D detection tasks. 
In the future, we would like to explore sparse padding and related interpolation operations in the following directions.

\paragraph{N-dimensional sparse data} Currently, our study is mainly performed in 3D, while all the sparse padding schemes are generalizable to any dimension. 
However, the increased runtime memory would be a severe side effect in high dimensions.

\paragraph{Non-regular interpolation scheme} As the increased feature maps from the padded voxels are the main memory bottleneck, it would be interesting to employ non-regular interpolation schemes like RBF (radial basis function interpolation) to avoid pad empty voxels while enjoying the flexibility brought by feature interpolation.  The parameters of RBF are possibly learned during the training.


{\small
\bibliographystyle{ieee_fullname}
\bibliography{src/reference}
}



\section*{Supplementary material}

# Supplementary Material: Interpolation-Aware Padding for 3D Sparse Convolutional Neural Networks

Yu-Qi Yang[1,2]   Peng-Shuai Wang[2]   Yang Liu[2]
[1]Tsinghua University   [2]Microsoft Research Asia
yangyq18@mails.tsinghua.edu.cn   {penwan,yangliu}@microsoft.com

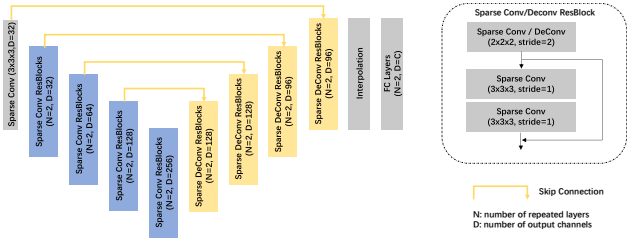

Figure 1: Network structure for PartNet Segmentation and Semantic KITTI. The numbers of repeated resblocks are [2,2,2,2,2,2,2,2].

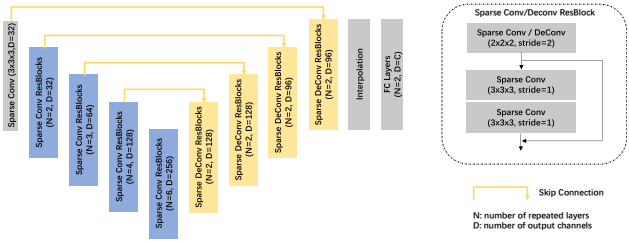

Figure 2: Network structure for Scannet segmentation. The numbers of repeated resblocks are [2,3,4,6,2,2,2,2].

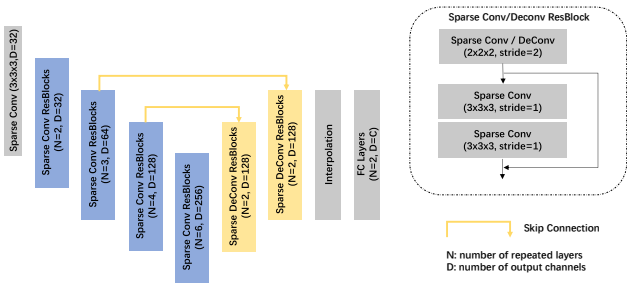

Figure 3: Network structure for Scannet segmentation when $s_{out} = 8\,\text{cm}$. The numbers of repeated resblocks are [2,3,4,6,2,2].

## 1. Detailed Segmentation Results on PartNet

The part IOUs of all categories are listed in Tab. 1. The results shown in the table are the average metrics and mean deviations of three rounds. On each round, all the networks are initialized with the same parameters.

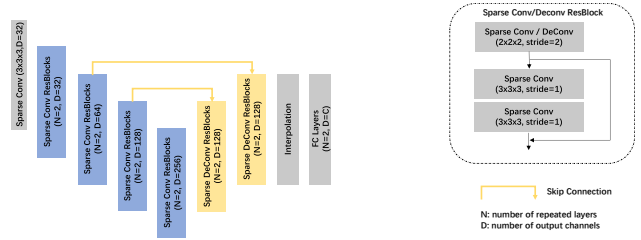

Figure 4: Network structure for Semantic KITTI when $s_{out} = 20\,\text{cm}$. It is also the network for Scannet detection, $s_{out} = 8\,\text{cm}$. The numbers of repeated resblocks are [2,2,2,2,2,2].

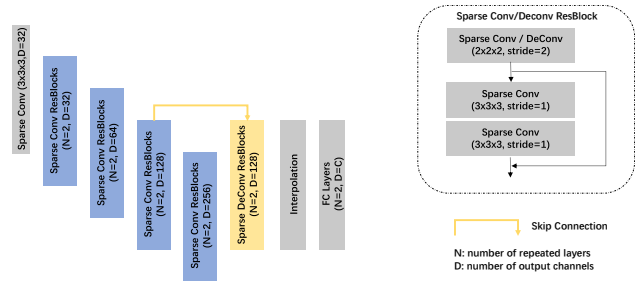

Figure 5: Network structure for Semantic KITTI when $s_{out} = 40\,\text{cm}$. The numbers of repeated resblocks are [2,2,2,2,2].

## 2. 3D Object Detection with a Stronger Baseline

H3DNet [5] is one of the state-of-the-art methods on 3D object detection of ScanNet. It uses the PointNet++[3] as the backbone to extract point features and predicts geometric primitives with these features, then generates object proposals. The results of H3DNet when using one PointNet++ is reported in Tab. 2. After replacing the PointNet++ by a sparse-voxel-based U-Net with our interpolation-aware sparse padding and the trilinear interpolation, we can achieve 47.1mAP@0.5, and the performance gain compared with H3DNet is 3.4. We also observe that compared with our own results based on VoteNet (see the 3rd row in Tab. 2), the performance is also greatly improved, which indicates that

| Group | Pad. | Interp. | $S_{out}$ | mIOU | Chair | Lamp | Stora | Table |
|---|---|---|---|---|---|---|---|---|
| (1) | ZERO | NEAR | $64^3$ | $40.5 \pm 0.2$ | $46.2 \pm 0.2$ | $24.2 \pm 0.3$ | $53.7 \pm 0.3$ | $37.8 \pm 0.4$ |
| | OCTREE | NEAR | $64^3$ | $40.0 \pm 0.0$ | $46.3 \pm 0.2$ | $23.7 \pm 0.1$ | $53.4 \pm 0.2$ | $36.7 \pm 0.2$ |
| | RING | NEAR | $64^3$ | $\mathbf{40.9} \pm 0.0$ | $\mathbf{47.3} \pm 0.2$ | $\mathbf{24.4} \pm 0.1$ | $\mathbf{53.8} \pm 0.1$ | $\mathbf{38.2} \pm 0.1$ |
| | INTERP | NEAR | $64^3$ | $40.6 \pm 0.1$ | $46.9 \pm 0.2$ | $24.1 \pm 0.2$ | $53.4 \pm 0.2$ | $37.9 \pm 0.5$ |
| (2) | ZERO | LINEAR | $64^3$ | $41.5 \pm 0.0$ | $46.8 \pm 0.3$ | $28.0 \pm 0.1$ | $53.3 \pm 0.5$ | $37.9 \pm 0.1$ |
| | OCTREE | LINEAR | $64^3$ | $41.4 \pm 0.0$ | $47.3 \pm 0.3$ | $27.2 \pm 0.4$ | $53.6 \pm 0.5$ | $37.6 \pm 0.1$ |
| | RING | LINEAR | $64^3$ | $\mathbf{42.7} \pm 0.3$ | $\mathbf{48.0} \pm 0.4$ | $\mathbf{28.7} \pm 0.4$ | $\mathbf{54.4} \pm 0.2$ | $\mathbf{39.7} \pm 0.4$ |
| | INTERP | LINEAR | $64^3$ | $42.3 \pm 0.3$ | $47.6 \pm 0.2$ | $28.6 \pm 0.5$ | $54.3 \pm 0.1$ | $38.8 \pm 0.5$ |
| (3) | ZERO | NEAR | $32^3$ | $38.1 \pm 0.2$ | $45.1 \pm 0.2$ | $23.0 \pm 0.4$ | $47.9 \pm 0.1$ | $36.3 \pm 0.2$ |
| | INTERP | LINEAR | $32^3$ | $\mathbf{40.1} \pm 0.1$ | $46.2 \pm 0.1$ | $27.3 \pm 0.3$ | $49.7 \pm 0.5$ | $37.1 \pm 0.2$ |

Table 1: Quality statistics of fine-grained part segmentation on four PartNet categories. *Pad.* is the sparse padding type, *Int.* is the sparse interpolation type, *mIoU* is the average part IoU

| Network | Pad. | Int | mAP@0.25 | mAP@0.5 |
|---|---|---|---|---|
| VoteNet [2] | - | - | $57.8 \pm 0.6$ | $34.7 \pm 0.4$ |
| MinkNet [1] | ZERO | NEAR | $58.7 \pm 0.5$ | $37.9 \pm 0.6$ |
| Ours | INTERP | LINEAR | $\mathbf{60.7} \pm 0.8$ | $\mathbf{41.4} \pm 0.6$ |
| H3DNet [5] | - | - | $64.0 \pm 0.7$ | $44.7 \pm 0.8$ |
| MinkNet [1] | ZERO | NEAR | $63.6 \pm 0.4$ | $45.8 \pm 0.4$ |
| Ours | INTERP | LINEAR | $\mathbf{64.4} \pm 0.2$ | $\mathbf{47.1} \pm 0.3$ |

Table 2: Quality statistics of instance detection on the ScanNet validation set.

with a stronger baseline, we can achieve get better results.

## 3. Network Structures

### 3.1. PartNet Segmentation

We use the U-Net structure with five levels of domain resolution. The finest grid resolution in the network is set to $64^3$. The structure is shown in Fig. 1. The numbers of repeated resblocks are [2,2,2,2,2,2,2,2].

### 3.2. ScanNet Segmentation

For the ScanNet Segmentation task, We use the same U-Net structure in [1] with five levels of domain resolution. The finest grid resolution in the network is set to $2\,\text{cm}$. The structure is shown in Fig. 2. The numbers of repeated resblocks are [2,3,4,6,2,2,2,2]. When the $s_{out}$ is set to $8\,\text{cm}$, we remove two high resolution layers in the decoder, and interpolate the feature from the 3rd level. The network structure is shown in Fig. 3.

### 3.3. Semantic KITTI Segmentation

For Semantic KITTI Segmentation, we use the same U-Net structure as [4]. It is same with the network shown in Fig.1. The finest grid resolution in the network is set to $5\,\text{cm}$. When the $s_{out}$ is set to $20\,\text{cm}$ or $40\,\text{cm}$, we also remove the corresponding layers in the decoder, the network structures are shown in Fig. 4 and Fig. 5.

### 3.4. ScanNet Detection

For the ScanNet detection task, we follow the pipeline of VoteNet [2], and only replace the PointNet++ with U-Net based on the sparse-convolution. The finest grid resolution in the network is set to $2\,\text{cm}$. And the $s_{out}$ is fixed as $8\,\text{cm}$ in this task. The U-Net structure is the same as the network shown in Fig. 5. The numbers of repeated resblocks are [2,2,2,2,2,2].

## 4. Advantage with 1-ring padding

Though the performance of the 1-ring padding is slightly better than the interpolation-aware padding on PartNet, its memory and computational cost are much higher, especially on large-scale datasets such as ScanNet and KITTI. The comparison of memory consumption with batch size as 1 is summarized in Table 3. In practice, we train the network with a batch size of at least 4 due to the requirement of Batch Normalization, otherwise, the performance may decrease greatly. And if the 1-ring padding is used, the network runs out of memory even on V100 GPU with 32GB memory on ScanNet and KITTI, despite its potential performance improvement. Our interpolation-aware padding is more practical and provides clear improvements over previous approaches.

| Padding scheme | ScanNet | KITTI |
|---|---|---|
| 1-ring padding | 8.9G | 12.9G |
| Interpolation-aware padding | 6.7G | 6.4G |

Table 3: The memory cost comparison with batch size 1.

## 5. Source code

Our source code is available on our project homepage: *https://yukichiii.github.io/project/padding.html*, feel free to follow the instruction in ReadMe to use our padding scheme.



\end{document}